\documentclass[a4paper,11pt,oneside,onecolumn]{article}

\usepackage[utf8]{inputenc}
\usepackage{fancyhdr}
\usepackage{authblk}
\usepackage[a4paper]{geometry}

\usepackage[pdftex,colorlinks=true,urlcolor=blue,citecolor=black,anchorcolor=black,linkcolor=black]{hyperref}

\usepackage{cite}
\usepackage{amsmath,amssymb,amsfonts}
\usepackage{algorithmic}
\usepackage{graphicx}
\usepackage{textcomp}
\usepackage{xcolor}
\usepackage{booktabs}
\usepackage{makecell}
\usepackage{flushend}

\def\BibTeX{{\rm B\kern-.05em{\sc i\kern-.025em b}\kern-.08em
    T\kern-.1667em\lower.7ex\hbox{E}\kern-.125emX}}

\usepackage[acronym, nomain, nowarn]{glossaries}
\makeglossaries
\newacronym{ai}{AI}{Artificial Intelligence}
\newacronym{arima}{ARIMA}{Autoregressive Integrated Moving Average}
\newacronym{slp}{SLPs}{Synthesized Load Profiles}
\newacronym[plural=LSTMs,firstplural=Long short-term memory models]{lstm}{LSTM}{Long Short-Term Memory}
\newacronym[plural=LLMs,firstplural=Large Language Models]{llm}{LLM}{Large Language Models}
\newacronym[plural=xLSTMs, firstplural=extended LSTMs]{xlstm}{xLSTM}{Extended Long Short-Term Memory}
\newacronym{nlp}{NLP}{Natural Language Processing}
\newacronym[plural=FCNs, firstplural=Fully-connected Networks]{fcn}{FCN}{Fully-connected Network}
\newacronym[plural=ANNs, firstplural=Artificial Neural Networks]{ann}{ANN}{Artificial Neural Network}
\newacronym{ml}{ML}{Machine Learning}
\newacronym{ar}{AR}{autoregressive}
\newacronym{tft}{TFT}{Temporal Fusion Transformer}
\newacronym{sarima}{SARIMA}{seasonal ARIMA}
\newacronym{arimax}{ARIMAX}{ARIMA with inclusion of eXogenous variables}
\newacronym{relu}{RelU}{Rectified Linear Unit}
\newacronym{adam}{Adam}{Adaptive Moment Estimation}
\newacronym[plural=RNNs, firstplural=Recurrent Neural Networks]{rnn}{RNN}{Recurrent Neural Network}
\newacronym{gru}{GRU}{Gated Recurrent Unit}
\newacronym{slstm}{sLSTM}{scalar LSTM}
\newacronym{mlstm}{mLSTM}{matrix LSTM}
\newacronym{mse}{MSE}{Mean Squared Error}
\newacronym[plural=RMSEs, firstplural=Root Mean Squared Errors]{rmse}{RMSE}{Root Mean Squared Error}
\newacronym{nrmse}{nRMSE}{normalized RMSE}
\newacronym{mae}{MAE}{Mean Absolute Error}
\newacronym{mape}{MAPE}{Mean Absolute Percentage Error}
\newacronym{gpt}{GPT}{Generative Pre-trained Transformer}
\newacronym{bert}{BERT}{Bidirectional Encoder Representations from Transformers}
\newacronym[plural=SVMs, firstplural=Support Vector Machines]{svm}{SVM}{Support Vector Machine}
\newacronym{gpr}{GPR}{Gaussian Process Regression}
\newacronym{emd}{EMD}{Empirical Mode Decomposition}
\newacronym{ica}{ICA}{Imperialistic Competitive Algorithm}
\newacronym{mane}{MANE}{Mean Absolute Normalized Error}
\newacronym{hvac}{HVAC}{heat, ventilation, air, conditioning}
\newacronym{mlp}{MLP}{Multi-Layer Perceptron}
\newacronym{svr}{SVR}{Support Vector Regression}
\newacronym{dnn}{DNN}{Deep Neural Network}
\newacronym[plural=NNs, first=Neural Networks]{nn}{NN}{Neural Network}
\newacronym{xgboost}{XGBoost}{eXtreme Gradient Boosting}
\newacronym[plural=GJs, firstplural=Giga-Joules]{gj}{GJ}{Giga-Joule}
\newacronym{dhc}{DHC}{District Heating and Cooling}
\newacronym{arma}{ARMA}{Autoregressive Moving Average}
\newacronym{armax}{ARMAX}{ARMA with eXogenous inputs}
\newacronym{cnn}{CNN}{Convolutional Neural Network}
\newacronym[plural=SSMs, firstplural=State Space Models]{ssm}{SSM}{State Space Model}
\newacronym{smape}{SMAPE}{Symmetric MAPE}
\newacronym{te}{TE}{Transformer-Encoder}
\newacronym[plural=GRNs, firstplural=Gated Residual Networks]{grn}{GRN}{Gated Residual Network}
\newacronym[plural=GLUs, first=Gated Linear Units]{glu}{GLU}{Gated Linear Unit}
\newacronym[plural=VSNs, first=Variable Selection Networks]{vsn}{VSN}{Variable Selection Network}
\newacronym{iqr}{IQR}{Interquartile Range}
\newacronym{lod}{LoD}{Level of Detail}
\newacronym{rse}{RSE}{Rooted Squared Error}
\newacronym{nrse}{nRSE}{normalized RSE}
\newacronym[plural=CPUs, firstplural=Central Processing Units]{cpu}{CPU}{Central Processing Unit}
\newacronym[plural=GPUs, firstplural=General Processing Units]{gpu}{GPU}{Graphics Processing Unit}
\newacronym[plural=FMs, first=Foundation Models]{fm}{FM}{Foundation Model}
\newacronym[plural=kWh, first=kilowatt hours]{kwh}{kWh}{kilowatt hours}
\newacronym{chp}{CHP}{combined heat and power}

\usepackage[nolist]{acronym}
\begin{acronym}[XXXXX]
	\acro{COP}{coefficient of performance}
	\acro{BESS}{Battery Energy Storage System}
	\acro{DT}{Digital Twin}
	\acro{DHW}{domestic hot water}
	\acro{HDD}{heating degree day}
	\acro{HEMS}{Home Energy Management System}
	\acro{EU}{European Union}
	\acro{EV}{Electric Vehicle}
	\acro{HP}{Heat Pump}
	\acro{HVAC}{heating, ventilation and air-conditioning}
    \acro{IQR}{Interquartile Range}
	\acro{LP}{Linear Program}
    \acro{ML}{Machine Learning}
    \acro{MPC}{Model Predictive Control}
	\acro{NPV}{Net Present Value}
    \acro{PV}{Photovoltaic}
	\acro{SPF}{Seasonal Performance Factor}
\end{acronym}

\fancypagestyle{firstpagefooter}{%
  \fancyhf{}
  
  \fancyfoot[L]{Submitted version of the paper submitted to IEEE SusTech, 2026\\
  DOI: not yet available\\
  \copyright~2026 IEEE}
}

\title{Benchmarking Transformer and xLSTM for Time-Series Forecasting of Heat Consumption}

\author[1]{Marja Wahl\footnote{Email address: marja.wahl@rausch.se}}
\author[2]{Daniel R. Bayer\footnote{Email address: daniel.bayer@uni-wuerzburg.de}}
\author[1]{Sven Rausch}
\author[2]{Marco Pruckner}
\affil[1]{RAUSCH Technology GmbH, W\"urzburg, Germany}
\affil[2]{Modeling and Simulation, University of W\"urzburg, W\"urzburg, Germany}
\date{}

\begin{document}

\maketitle

\begin{abstract}
    Obtaining an accurate short-term forecasting for heat demand is an essential part of operating district heating networks cost-efficient and reliable.
    Heat consumption time series at the building level are highly dependent on exogenous variables such as outdoor temperature and individual usage patterns, making forecasting in this context a challenging task.
    Thus, this paper benchmarks novel Transformer-based and xLSTM architectures for short-term heat-demand forecasting.
    Using hourly data from 25 German buildings (2017–2025), we compare three-hour and 24-hour forecasting horizons relevant for intraday control and day-ahead scheduling.
    We establish a multi-building benchmark that tests whether models trained on pooled, heterogeneous building data are able to generalize across diverse building stock.
    The results show that the xLSTM achieves the lowest RMSE (19.88~kWh for three-hour, 21.47~kWh for 24-hour forecasts), while the Temporal Fusion Transformer attains the best MAE (9.16~kWh for three-hour forecasts).

    As xLSTMs and Transformers require long training times and have a huge number of trainable parameters, their sustainability remains questionable.
    Therefore, this paper further investigates the trade-off between predictive accuracy and computational resource demand of the evaluated forecasting models.
    The findings indicate that also low-parameter models like a traditional fully-connected network achieve good predictive results, highlighting that marginal accuracy gains of the novel prediction models come at substantial resource expense for this use case.
    %
\end{abstract}

\thispagestyle{firstpagefooter}


\section{Introduction}

The energy sector accounts for about 75~\% of global greenhouse gas emissions \cite{energy_and_climate}, roughly 50~\% of those are attributable to the heating sector alone \cite{energy_transition}.
This makes the heating sector a major lever for decarbonization.
In the special case of district heating systems, multiple generation assets (e.g., combined heat and power units, boilers, large heat pumps, and thermal storage) must be scheduled efficiently.
While an overgeneration of heat increases both carbon emissions and operational costs, insufficient heat supply leads to thermal discomfort and complaints of residents connected to a district heating network.
Therefore, accurate heat demand forecasts are essential to ensure sustainable and user-friendly operation.
While our dataset and partners are based in Germany, district heating is an important pillar for sustainable heating worldwide, and forecasting advances translate broadly to urban energy systems. When reliable short-term forecasts are available, data-driven optimization yields substantial energy savings and CO$_2$ reductions \cite{djebko_design_2024, EnergyInformatics2020, CarbonPeak2021}; \gls{ml} models further improve predictive accuracy on energy time series \cite{Feng2024}.

Classical statistical approaches (e.g., ARIMA or simple linear predictors) struggle with heterogeneous, non-stationary, and multivariate demand signals \cite{zhu2023time, bayer2024electricity}. Deep neural architectures try to mitigate these issues by learning non-linear dynamics from large datasets. Examples like \glspl{lstm} alleviate vanishing gradients but remain sequential and can degrade on long contexts \cite{sherstinsky_fundamentals_2023}. Transformers replace recurrence with self-attention, enabling parallel training and long-range interactions \cite{vaswani_attention_nodate, zhou_informer_2021}. Recently, the \gls{xlstm} aims to combine gated recurrence with improved long-range memory and partial parallelism \cite{beck2024xlstmextendedlongshortterm, tay2021long}. However, many prior studies evaluate at an aggregate (network/district) level or on very few buildings; it remains unclear whether modern models benefit from training on data pooled across many buildings and how well they generalize under real heterogeneity.

This paper benchmarks Transformer-based and xLSTM-based architectures for short-term heat-demand forecasting under realistic, sustainability-aware constraints. We study three-hour and 24-hour horizons, directly supporting intraday control and day-ahead scheduling.

In detail, the contributions of this paper are threefold:
\begin{enumerate}
    \item A multi-building benchmark for heat demand forecasting is established, including measurements from 25 German (residential and commercial) buildings spanning 2017--2025 with hourly resolutions and heterogeneous static attributes;
    \item A head-to-head comparison of modern models like \gls{xlstm} and Transformer families against strong baselines (\gls{fcn}, \gls{lstm}) is performed, using standardized inputs (history, weather, calendar, building features) and reporting multiple error metrics;
    \item A computational sustainability assessment is conducted, analyzing training time and memory/parameter footprint to link model choice to operational carbon and feasibility for edge deployment. 
\end{enumerate}

By targeting forecasting needs that directly influence dispatch, fuel use, and emissions—and by coupling predictive accuracy with computational efficiency—this study advances intelligent, low-carbon heat operations.

The remainder of this work is organized as follows: Section \ref{sec:relatedwork} reviews existing literature, Section~\ref{sec:method} describes our methodology and evaluation approach, Section~\ref{sec:results} presents experimental results for different forecasting horizons, Section~\ref{sec:discussion} discusses implications and comparisons with related work, and Section \ref{sec:conclusion} concludes the paper.

\section{Related Work}
\label{sec:relatedwork}

Energy forecasting is a widely studied field covering all sectors, including electricity, oil, gas, and district heating consumption, with many works focusing on electricity data and fewer on heat consumption \cite{hong_energy_2020}.

In the concrete case of heat demand forecasting, different algorithmic approaches are used.
For example, Eseye et al. \cite{eseye_short-term_2020} forecast 24-hour heat demand of four buildings using \glspl{svm} with calendar, weather, and historical load features, enhanced by empirical mode decomposition and a genetic algorithm for feature selection, achieving a \gls{mape} of 4.65~\%.
Another traditional \gls{ml} model, namely \gls{gpr}, is used by Potočnik et al. \cite{potocnik_machine-learning-based_2021} for 48-hour heat demand forecasting on city-wide data, outperforming \glspl{nn} and linear models with a \gls{mane} of 2.94~\%, while highlighting inherent forecasting errors from unknown heating processes.
As a result, their analysis confirms that accurate temperature forecasts increase the quality of heat demand forecasting using traditional \gls{ml} models, while other weather forecasts are not that important.
A further example for the combination of manual preprocessing and \gls{ml} methods like done by Eseye et al. \cite{eseye_short-term_2020} is presented by Tang et al. \cite{tang_modeling_2014}.
Using a clustering approach in combination with an ensemble of \gls{mlp} models, they are able to  achieve a \gls{mape} of 8.63~\% for HVAC demand forecasting.

In the literature, also \gls{nn}-based approaches are commonly used for heat demand prediction.
For instance, Kato et al. \cite{kato_heat_2008} implement recurrent and non-recurrent \glspl{nn} for district heat demand prediction, noting that the \gls{rnn} outperforms feed-forward networks on short-term horizons.
When comparing recurrent networks with statistical methods, Muzaffar et al. \cite{muzaffar_short-term_2019} present for the use case of electricity load forecasting that \glspl{lstm} outperform statistical methods for short-term forecasts up to two days.
Focusing specifically on these \glspl{lstm}, Li et al. \cite{li_assessment_2021} compare pure and hybrid \gls{lstm} models for predicting building electricity consumption, finding simple \gls{lstm} variants often outperform hybrids.

A comparison of different algorithmic approaches in the case of district heat demand prediction is given for instance by Xue et al. \cite{xue_multi-step_2019}, where the performance of a \gls{svr}, deep \glspl{nn}, and \gls{xgboost} is evaluated for 24-hour heat load forecasting.
They report that recursive \gls{xgboost} achieves a \gls{mape} of 9.59~\%.
A comparison to recurrent networks was not presented, thus limiting comparability to other works like \cite{muzaffar_short-term_2019} or \cite{li_assessment_2021}.

In addition to \gls{rnn}, the Transformer developed as a novel architecture.
Transformers require positional encodings to handle sequence order, implemented as fixed or learnable embeddings \cite{vaswani_attention_nodate, zerveas_transformer-based_2020}.
To reduce self-attention's quadratic complexity, models like Informer \cite{zhou_informer_2021} and Autoformer \cite{wu_autoformer_2021} use sparse attention or hierarchical architectures.
These Transformers show strong performance in energy forecasting for electricity demand and generation.
For instance, Zhao et al. \cite{zhao_short-term_2021} use a vanilla Transformer with learnable embeddings and similar day selection for day-ahead electricity load, outperforming \gls{lstm} and GRU.
Zhu et al. \cite{zhu2023time} propose SL-Transformer for wind and solar power forecasting, where the authors claim that it outperforms other approaches.
To further reduce the \gls{mape} on electricity load prediction, Zhang et al. \cite{zhang_short-term_2022} introduce a Time Augmented Transformer with a temporal embedding layer.
In the recent past, \glspl{tft} have demonstrated very good prediction qualities in the field of  electricity forecasting \cite{huy_short-term_2022, giacomazzi2023short}.

Besides transformer-based approaches, the \gls{lstm} architecture has recently been extended by multiple memory cells to a \gls{xlstm}, improving long-term pattern storage \cite{beck2024xlstmextendedlongshortterm}.
Adaptations for time series like xLSTMTime \cite{alharthi2024xlstmtime} and xLSTM-Mixer \cite{kraus_xlstm-mixer_2024} incorporate normalization and linear layers, outperforming several Transformer models on benchmark datasets.

\subsection{Open Research Questions and Contribution}
Heat demand data is highly auto-correlated, but also strongly influenced by exogenous variables like the ambient temperature.
Thus, the question remains how good the novel \gls{nn} architectures like the \gls{xlstm} or the Transformer perform for such time series compared to traditional approaches.
Furthermore, while most prior works train separate models per building or one single aggregated district heating network instance, this work benchmarks various models for heat load forecasting across multiple buildings and horizons to test generalization capabilities, assessing both accuracy and training effort.
Unlike other works, it includes residential data with gas and district heat consumption, an area less studied compared to commercial or industrial buildings.
Finally, as these novel architectures generally require higher computational resources during training, their predictive performance is analyzed relative to their training cost.


\section{Method}\label{sec:method}

%
%

This section describes the methodology used for benchmarking Transformer and \gls{xlstm} architectures on heat data forecasting. We provide a brief overview of the model architectures used in this study.

\subsection{Model Architectures}

\subsubsection{Fully Connected Network (FCN)}
As a simple baseline model, we implement a \gls{fcn} with one hidden layer. \glspl{fcn} are straightforward \glspl{nn} that consist of fully connected layers where each neuron is connected to every neuron in the adjacent layers. Despite their simplicity, \glspl{fcn} can serve as effective baselines for time-series forecasting tasks. Our implementation uses a single hidden layer with ReLU activation function, followed by a dropout layer, and finally, a linear output layer.

\subsubsection{Long Short-Term Memory (LSTM)}
The \gls{lstm} network~\cite{hochreiter1997long} serves as our second baseline model. \glspl{lstm} are \glspl{rnn} specifically designed to address the vanishing gradient problem in traditional \glspl{rnn}, making them well-suited for sequence modeling tasks. The \gls{lstm} architecture includes memory cells with input, output, and forget gates that control information flow. 
The implementation of the \gls{lstm} in our experiments is equal to that of the \gls{fcn}, with the hidden layer being a \gls{lstm} layer of variable size and the activation function set to tanh.

\subsubsection{xLSTM}

The extended Long Short-Term Memory (xLSTM) model~\cite{beck2024xlstmextendedlongshortterm} addresses several limitations of traditional \glspl{lstm}, including the inability to revise storage decisions, limited storage capabilities, and lack of parallelization. The \gls{xlstm} architecture achieves strong performance on long-range sequence tasks, making it particularly suitable for time-series forecasting.

\begin{figure}[htbp]
    \centering
    \makebox[\columnwidth][c]{\includegraphics[width=\columnwidth]{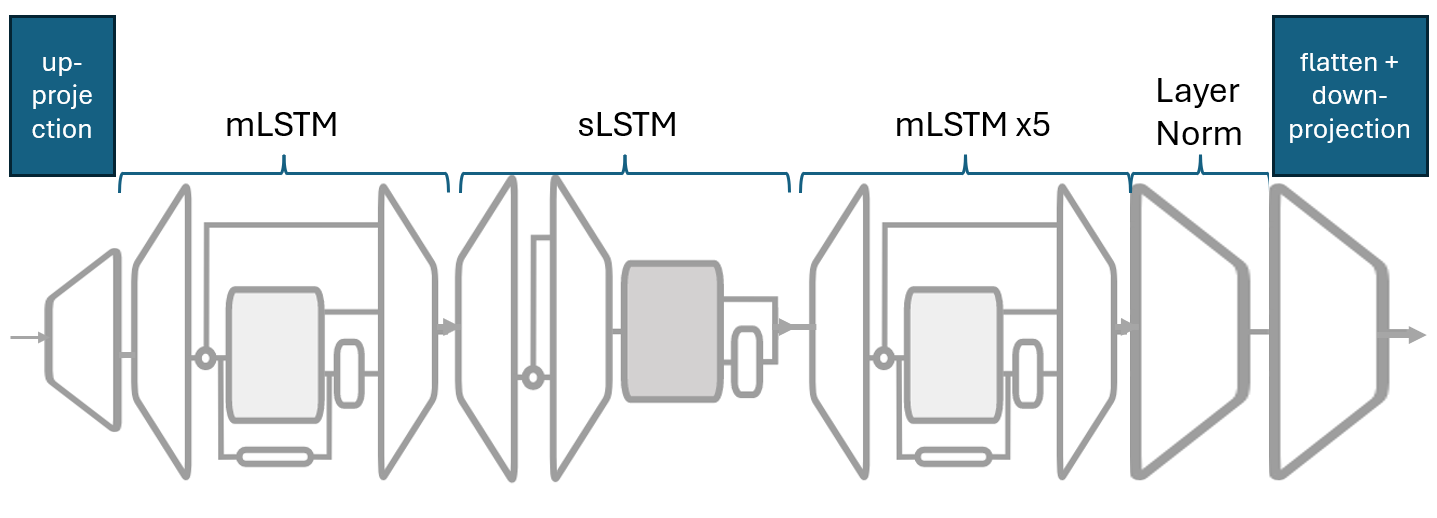}}
    \caption{Architecture of the adapted \gls{xlstm}. mLSTM and sLSTM are ordered as in the
original \gls{xlstm} paper \cite{beck2024xlstmextendedlongshortterm}, added layers are described with blue text box.}
    \label{fig:xlstm-architecture}
\end{figure}

Our implementation follows the architecture proposed by Beck et al.~\cite{beck2024xlstmextendedlongshortterm}, with minor adaptations for heat data forecasting. Figure~\ref{fig:xlstm-architecture} displays the original \gls{xlstm} and the added layers. First, the input is passed into an up-projection layer, which increases the third dimension of the input data to hidden dimension size $d=256$. The output is then fed to the \gls{xlstm} blocks, mLSTM and sLSTM. For a further description, we refer to the original paper by Beck et al. \cite{beck2024xlstmextendedlongshortterm}. The number of attention heads and the size of the 1D-convolutions is set to four. The context length is set to the input size $n_{in}$, and for the mLSTM layers, the query-key-value size is set to four. A post-normalization layer of size $d$ is added after, as in the original implementation. Finally, the vector is flattened to length $d \cdot n_{in}$ and down-projected to output length $n_{out}$.

\subsubsection{Transformer Encoder (TE)}
We implement a Transformer-based architecture following the attention mechanism introduced by Vaswani et al.~\cite{vaswani_attention_nodate}. Transformer models are attractive for time-series forecasting due to their ability to process data in parallel and focus on important dates in the time-series data through the attention mechanism. This parallelization allows Transformer models to outpace traditional \glspl{lstm} when training on large datasets.

In our experiments, we only test the \gls{te}, instead of opting for a full encoder-decoder architecture. This ensures that the model stays smaller in size, considering that the Encoder is able to generate an encoding of the input time series that is expressive enough. In our implementation, we use an initial up-projection layer and a learnable positional embedding layer, followed by four Encoder blocks, whose output is passed through a global average pooling, feed forward, and dropout layer.

\subsubsection{Temporal Fusion Transformer (TFT)}
We also evaluate the \gls{tft} model proposed by Lim et al.~\cite{lim2020temporalfusiontransformersinterpretable}, which is specifically designed for multi-horizon time-series forecasting. The \gls{tft} architecture combines \gls{lstm} layers with self-attention mechanisms to handle both temporal relationships and variable selection. The model includes specialized components for processing static covariates, known future inputs, and observed historical inputs, making it well-suited for heat load forecasting where both historical consumption patterns and future weather forecasts are relevant.

\subsection{Dataset and Preprocessing}
The dataset features hourly measured heat consumption in \gls{kwh}, obtained in the course of different research projects \cite{djebko_design_2024, ki_in_fernwaerme}. Overall, data from 25 different buildings, either running on natural gas or district heat energy, is collected. The time series are of different length and from different periods of the year, all gathered between 2017 and 2025.

The data undergoes several preprocessing steps before model training:

\begin{itemize}
\item \textbf{Data cleaning}: Removal of outliers due to connection errors using the \gls{iqr} method
\item \textbf{Standardization}: Per-series standardization to ensure scale invariance during training
\item \textbf{Splitting and Interpolating}: Series are split if they are missing data spans of longer than 24 hours and linearly interpolated if they are missing less
\item \textbf{Feature engineering}: Additional features are derived from calendar data (day of month, day of week, holiday), building characteristics (heated area, number of apartments, type of heating), weather information (10 different meteorological features provided by meteostat\footnote{https://dev.meteostat.net/python/hourly.html}), and the historic heat consumption itself (average daily heat consumption)
\end{itemize}

After data preprocessing and adding external features, the hourly dataset amounts to 136.043 data points from 44 different time series.

\subsection{Evaluation Metrics}

We evaluate model performance using several metrics, including the \gls{mse} $ = \frac{1}{k} \sum_{t=1}^{k} (y_{t} - \hat{y}_{t})^2$, where $y_t$ is the actual value, $\hat{y}_t$ is the predicted value, and $k$ is the forecast horizon length. Further, to provide a comprehensive assessment of model performance, we evaluate on the root of the \gls{mse}, the \gls{rmse}, and the \gls{mae} $ = \frac{1}{k} \sum_{t=1}^{k} |y_{t} - \hat{y}_{t}|$.

Additionally, we report the scale-invariant \gls{nrmse} $ = \frac{\sqrt{\frac{1}{k}\sum_{t=1}^{k}(y_t-\hat{y}_t)^2}}{\max(y)-\min(y)}$.
Although the range‑based normalization of the \gls{nrmse} is sensitive to extreme values, this effect is mitigated by our preprocessing pipeline, which removes outliers using the \gls{iqr} method. In addition, standard normalization is applied per building, ensuring that differences in absolute consumption levels do not distort comparisons across heterogeneous buildings.

While \gls{mape} is commonly used in forecasting literature \cite{bourdeau_modeling_2019}, we do not use it in this study due to its limitations when actual values are zero or close to zero, which occurs in heat consumption data during warmer periods. In such cases, \gls{mape} can result in very large or undefined errors, making it unsuitable for evaluation.

\section{Experimental Setup}

The following section describes the experiments, while additional information can be found in the corresponding repository\footnote{https://github.com/marja-w/energy-forecast-benchmark}. It includes code, more detailed descriptions of experiments, hyperparameters, and results.

\subsection{Na$\ddot{\i}$ve Forecast}
For comparison purposes, we implemented a na$\ddot{\i}$ve forecast that uses a simple approach: it predicts heat load values for the next $x$ hours after timestep $t$: $(t+1, t+2, \dots, t+x)$, as heat load from the last $x$ hours until timestep $t$: $(t-x, t-(x-1), \dots, t-1, t)$. This is an adaptation of the hour-before method used as baseline by Djebko et al. \cite{djebko_design_2024}.

\subsubsection{Hyperparameter Tuning}

\begin{table}[htbp]
    \renewcommand{\arraystretch}{1.15}
    \centering
    \caption{Hyperparameter Settings for best performing Models (3-hour/24-hour prediction)}
    \label{tab:compact-hyperparameters}
    \begin{tabular}{lcccc}
        \toprule
        \textbf{Model} & \textbf{Batch Size} & \textbf{Dropout} & \textbf{Epochs} & \textbf{Neurons/Heads} \\
        \midrule
        \gls{fcn} & 100/104 & 0.05/0.19 & 34/78 & 51/131 \\
        \gls{lstm} & 100/32 & 0.1/0.1 & 28/78 & 121/131 \\
        \gls{te} & 76/32 & 0.17/0.1 & 10/10 & 8/4 \\
        \gls{tft} & 64/64 & 0.1/0.1 & 20/20 & 4/4 \\
        \gls{xlstm} & 32/64 & 0.1/0.1 & 10/30 & 4/4 \\
        \bottomrule
    \end{tabular}
\end{table}

All models are dependent on a considerable amount of hyperparameters, some of them shown in Table \ref{tab:compact-hyperparameters}. The evaluation results presented in Section \ref{sec:results} are achieved using those hyperparameter configurations. For \gls{fcn} and \gls{lstm}, we employed Bayesian optimization through Weights and Biases Sweep\footnote{https://docs.wandb.ai/guides/sweeps/}, additionally exploring length of input sequence (max. 72) and length of future variables (max. $n_{in}$). The $n_{in}$ for the best \gls{fcn} training was 72 for three-hour and 38 for 24-hour prediction, with no future variables. For \gls{lstm}, it was 59 and 38 for $n_{in}$, and length of 0 and 24 future variables for three-hour and 24-hour prediction. The \gls{fcn} was tested with 134 hyperparameter sets for three-hour prediction and with 60 for 24-hour prediction, the \gls{lstm} with 30 and 32, accordingly. For each hyperparameter search, we choose the Adam optimizer with default learning rate 0.001 \cite{kingma2014adam}, the Glorot weight initializer \cite{glorot2010understanding}, and the \gls{mse} as loss function.

Due to the significantly longer train durations of \gls{te}, \gls{tft}, and \gls{xlstm}, we manually selected and tested parameters, while setting the input feature lengths to 72 hours.

Additionally, each model was evaluated with three feature configurations: \textbf{(1)} only past consumption values \textbf{(2)} all features excluding building information \textbf{(3)} all available features. Almost all models trained best when being fed all features, but the \gls{fcn} and \gls{te}, which yielded the best results when being trained on only the past consumption values. A more detailed analysis for 24-hour prediction, can be found in Tab. \ref{tab:fc_comp_hourly_24}. For the TFT, we used all features, as it specifically designed for training on multivariate time series.

\begin{table}[htbp]
\centering
\caption{Comparison of 24-hour prediction average RMSE and MAE for different training features (1,2,3). Number of seeds are in brackets behind model names, best errors for each model are highlighted.}
\label{tab:fc_comp_hourly_24}
\begin{tabular}{lcccc}
\toprule
& FCN (5) & \gls{lstm} (5) & TE (2) & xLSTM (2) \\
\midrule
\multicolumn{5}{l}{\textbf{RMSE (kWh)} $\pm$ Std} \\
1 & \textbf{24.92} $\pm$ 0.03 & 26.08 $\pm$ 0.24 & \textbf{27.13} $\pm$ 0.58 & 69.86 $\pm$ 0.25 \\
2 & 29.81 $\pm$ 1.40 & 27.08 $\pm$ 1.30 & 30.05 $\pm$ 0.14 & 24.66 $\pm$ 0.71 \\
3 & 35.02 $\pm$ 7.06 & \textbf{25.08} $\pm$ 0.26 & 29.84 $\pm$ 0.72 & \textbf{21.67} $\pm$ 0.88 \\
\multicolumn{5}{l}{\textbf{MAE (kWh)} $\pm$ Std} \\
1 & \textbf{11.69} $\pm$ 0.19 & 12.04 $\pm$ 0.12 & \textbf{12.75} $\pm$ 0.72 & 45.10 $\pm$ 0.18 \\
2 & 15.11 $\pm$ 1.43 & 12.63 $\pm$ 0.92 & 14.09 $\pm$ 0.25 & 12.62 $\pm$ 0.83 \\
3 & 18.86 $\pm$ 5.39 & \textbf{11.31} $\pm$ 0.28 & 14.29 $\pm$ 0.38 & \textbf{11.77} $\pm$ 0.46 \\
\bottomrule
\end{tabular}
\end{table}

\subsection{Training Data Generation}
In the preprocessed dataset, each row holds consumption and feature information about time step $t$. This data is transformed into a series of input features $X_{t-n_{in}, t}$ and target values $y_{t+1, t+n_{out}+1}$, with look-back size $n_{in}$ and look-forward length $n_{out}$. The sliding window technique is used, which divides a continuous time series into overlapping segments that serve as input-output pairs for model training. For each position of the window, the data within becomes the input features $X$, while the data points immediately following become the target values $y$ to predict. We use a step size of one and perform experiments for $n_{out} = 3$ and $n_{out} = 24$.

For training, we have one target time series and number of $n_c$ continuous covariate time series. When transforming the data, we treat all series the same, resulting in number of $c = 1 + n_c$ input feature series $X_{i, n_{in}}^c$ and one target series $Y_{i, n_{out}}$. Covariate target series can be used as future covariates by some models, which represent weather forecasts, for example.
As input for the \gls{fcn}, we flatten all input feature series into a row vector. 
Therefore, the input of the \gls{fcn} is a matrix of size $X = (n, c \cdot n_{in} + n_s)$, with $n$ number of data points and $n_s$ number of static features. For all other models, the input matrix size is defined as $X = (n, n_{in}, 1 + n_c + n_s)$, where static covariate values are transformed to a series by repeating them $t$ times. The \gls{tft} is the only model that handles static and future covariates separately.

\subsection{Train, Validation, and Test Data Split}
The data is split along the time axis, using a 80/10/10 split for train, validation, and test data. Consequently, the models see each time series during training. Yet, since the time series in the dataset cover different periods of time, the models might train on different time of year and amount of data for each time series. The training data has a length of 105,274 data points, 10,016 data points for validation, and 10,028 data points for testing.

\section{Results}\label{sec:results}

%
\subsection{Hourly Heat Load Forecasting for Sustainable Energy Management}

The models were evaluated on hourly heat load data for forecasting horizons of three hours and 24 hours. These time frames are particularly relevant for sustainable energy management: three-hour forecasts enable grid operators to make very short-term decisions in the intraday market, reducing waste and unnecessary costs, while 24-hour predictions provide detailed next-day consumption patterns for more efficient resource planning.

\subsubsection{Three-Hour Forecasting Performance}

\begin{table}[htbp]
    \renewcommand{\arraystretch}{1.15}
	\centering
    \caption{Evaluation results for best models for three-hour prediction. RMSE and MAE are in kWh.}
	\label{tab:results-3-hour-prediction}
	\begin{tabular}{lcccc}
		\toprule
		\textbf{Model} & \textbf{nRMSE} & \textbf{RMSE} & \textbf{MSE} & \textbf{MAE}\\
		\midrule
		Na$\ddot{\i}$ve & 0.6981 & 35.6050 & 1,267.7481 & 15.6114\\

		\gls{fcn} & 0.6485 & 22.4679 & 504.9629 & 10.5686\\

		\gls{lstm} & 0.6628 & 22.7919 & 519.5272 & 10.3654\\

		\gls{te} & 0.6685 & 22.9671 & 527.4964 & 11.0534\\

		\gls{tft} & \textbf{0.5221} & 21.8016 & 475.5771 & \textbf{9.1612}\\

		\gls{xlstm}* & 0.5535 & \textbf{19.8792} & \textbf{395.2333} & 10.3044\\
		\bottomrule
        \multicolumn{5}{l}{{\footnotesize Models marked with * were trained on a GPU.}}
	\end{tabular}
\end{table}

Table \ref{tab:results-3-hour-prediction} presents the evaluation results for three-hour forecasting. The \gls{xlstm} model achieved the lowest \gls{rmse} (19.88 \gls{kwh}) and \gls{mse} (395.23), outperforming the second-best model (\gls{tft}) by 8.8~\% on the \gls{rmse} metric. Meanwhile, the \gls{tft} model demonstrated superior performance in terms of \gls{nrmse} (0.52) and \gls{mae} (9.16 kWh).

These results show that advanced models like \gls{xlstm} and \gls{tft} can improve short-term forecasting accuracy compared to traditional approaches.

\subsubsection{24-Hour Forecasting Performance}

For the 24-hour forecasting horizon (Table \ref{tab:results-24-hour-prediction}), the \gls{xlstm} model again achieved the lowest \gls{rmse} (21.47 \gls{kwh}) and \gls{mse} (461.11), outperforming the \gls{tft} model by approximately 11.4~\% on the \gls{rmse} metric. The \gls{tft} model achieved the lowest \gls{mae} (10.58 kWh). Interestingly, no model was able to beat the na$\ddot{\i}$ve forecast in terms of \gls{nrmse} (0.52), suggesting challenges when forecasting certain regions of the data range.

\begin{table}[htbp]
    \renewcommand{\arraystretch}{1.15}
	\centering
    \caption{Evaluation results for best models for 24-hour prediction. RMSE and MAE are in kWh.}
	\label{tab:results-24-hour-prediction}
	\begin{tabular}{lcccc}
		\toprule
		\textbf{Model} & \textbf{nRMSE} & \textbf{RMSE} & \textbf{MSE} & \textbf{MAE}\\
		\midrule
		Na$\ddot{\i}$ve & \textbf{0.5156} & 24.9731 & 623.6585 & 11.6490\\

		\gls{fcn} & 0.7405 & 24.8871 & 619.5878 & 11.4869\\

		\gls{lstm} & 0.7188 & 24.9085 & 620.5474 & 11.3159\\

		\gls{te}* & 0.9728 & 29.1964 & 852.5017 & 13.8039\\

		\gls{tft} & 0.5820 & 24.2263 & 587.6394 & \textbf{10.5753}\\

		\gls{xlstm}* & 0.5723 & \textbf{21.4712} & \textbf{461.1092} & 11.4327\\
		\bottomrule
        \multicolumn{5}{l}{{\footnotesize Models marked with * were trained on a GPU.}}
	\end{tabular}
\end{table}

 \begin{figure}[h]
    \centering
    \makebox[\columnwidth][c]{\includegraphics[width=0.95\columnwidth]{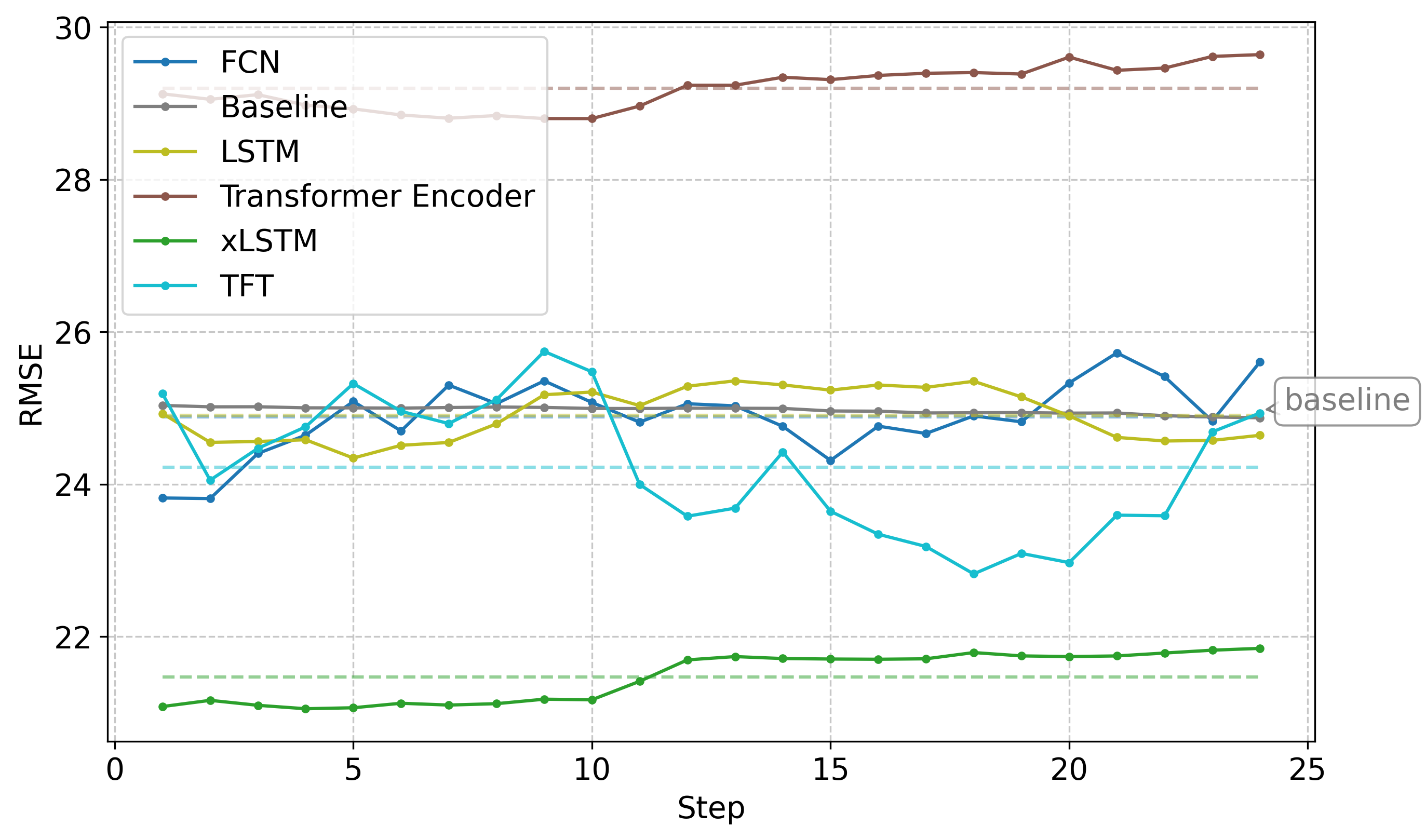}}
    \caption{Average \gls{rmse} (kWh) per forecasting step for each model in 24-hour prediction. Here, baseline refers to the na$\ddot{\i}$ve forecast.}
    \label{fig:per-step-24-hours}
\end{figure}

\begin{figure}[htbp]
    \centering
    \makebox[\columnwidth][c]{\includegraphics[width=0.95\columnwidth]{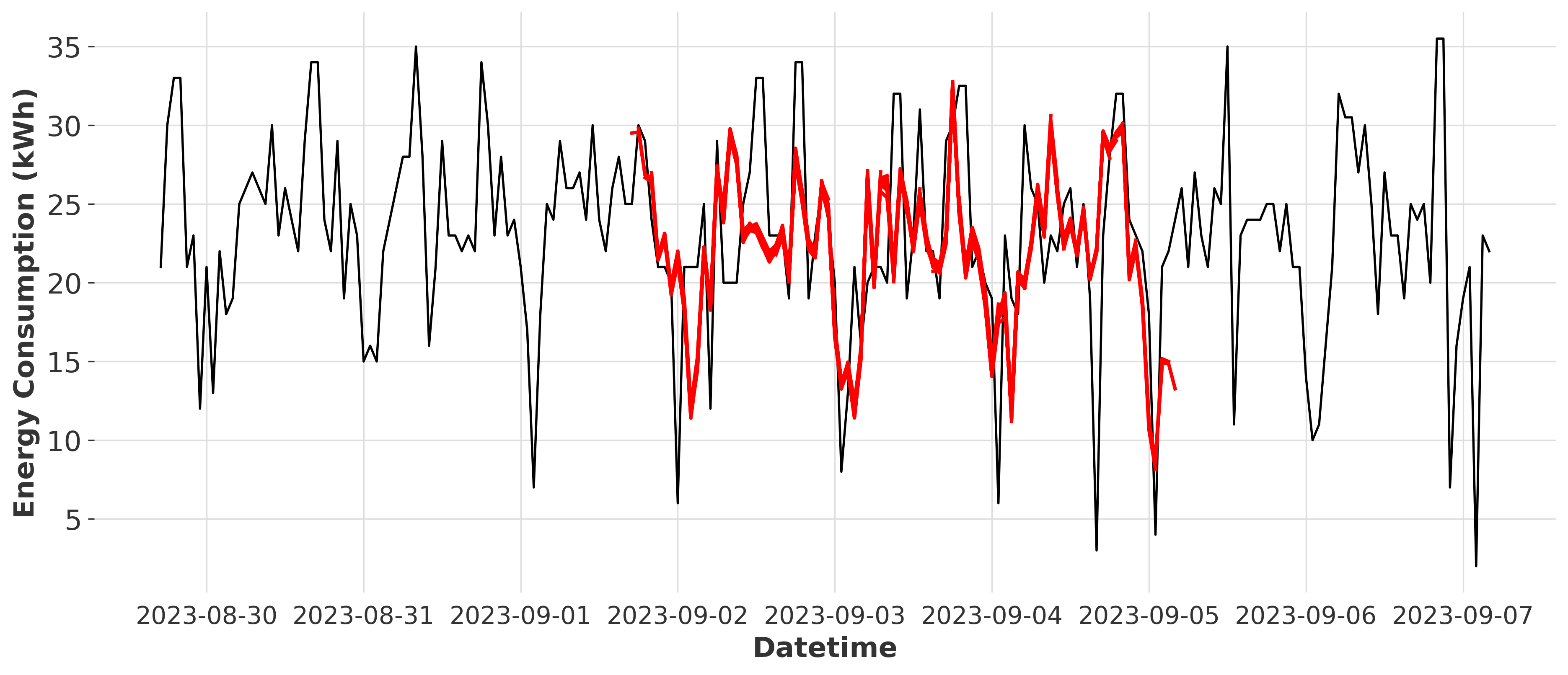}}
    \caption{Example of 24-hour forecasts of the \gls{xlstm} for one time series of the test set. The black line shows the recorded values, while the red line shows the 24-hour forecast.}
    \label{fig:xlstm-24-hour-forecast}
\end{figure}

Analysis of model performance across the 24-hour forecasting horizon (Fig. \ref{fig:per-step-24-hours}) revealed that \gls{xlstm} consistently outperformed other models at each forecasting step. The \gls{te} showed the weakest performance overall. The \gls{xlstm} maintained better-than-average performance until the eleventh forecasting step, suggesting its particular strength in medium-range predictions within the 24-hour window. An example forecast of the \gls{xlstm} can be seen in Fig. \ref{fig:xlstm-24-hour-forecast}.

\begin{figure}[h]
    \centering
    \includegraphics[width=0.90\columnwidth]{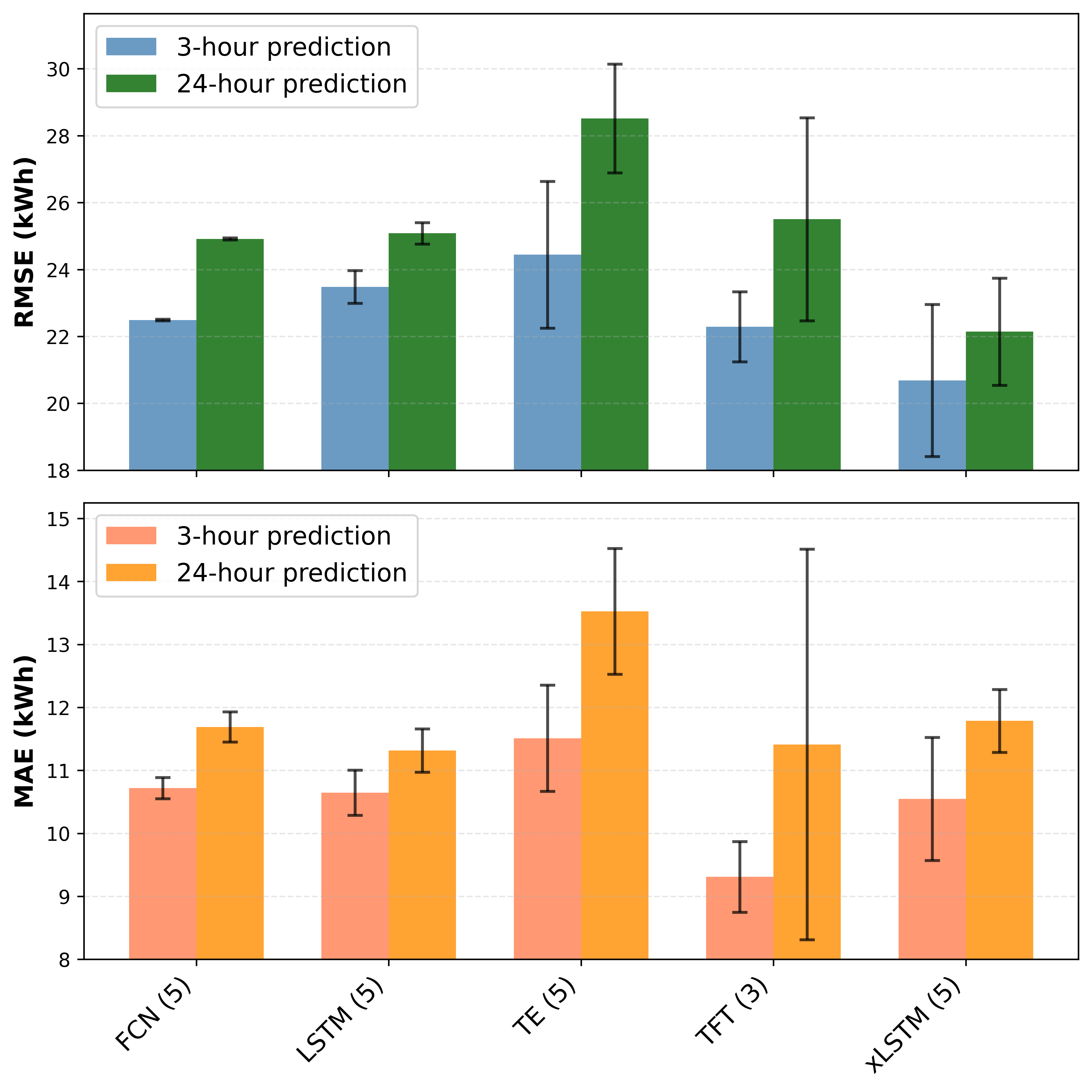}
    \caption{Comparison of average RMSE and MAE for 5-seeds of models with 95\% confidence intervals. Only the TFT has 3 seeds.}
    \label{fig:model-evaluation-with-confidence-intervals}
\end{figure}

\paragraph{Model Stability and Reliability}
We can analyze differences in stability of model training by averaging errors across training seeds, as presented in Fig. \ref{fig:model-evaluation-with-confidence-intervals}. The \gls{fcn} demonstrates the highest stability with the smallest confidence intervals, indicating highly consistent and reproducible results across training runs. In contrast, \gls{xlstm}, despite achieving the best overall accuracy, exhibits considerably higher variability, suggesting sensitivity to initialization and training conditions. \gls{tft} presents an interesting case with moderate variability for three-hour predictions but dramatically increased uncertainty for 24-hour forecasts, though the limited number of runs (3 vs. 5) makes definitive conclusions difficult. For sustainable energy applications, xLSTM's superior accuracy makes it the preferred choice for deployment, though practitioners should be prepared to invest more effort in the training and model selection process compared to the more stable but less accurate FCN.

\subsubsection{Building-Specific Performance}

\begin{figure}[htbp]
	\centering
	\makebox[\columnwidth][c]{\includegraphics[width=0.90\columnwidth]{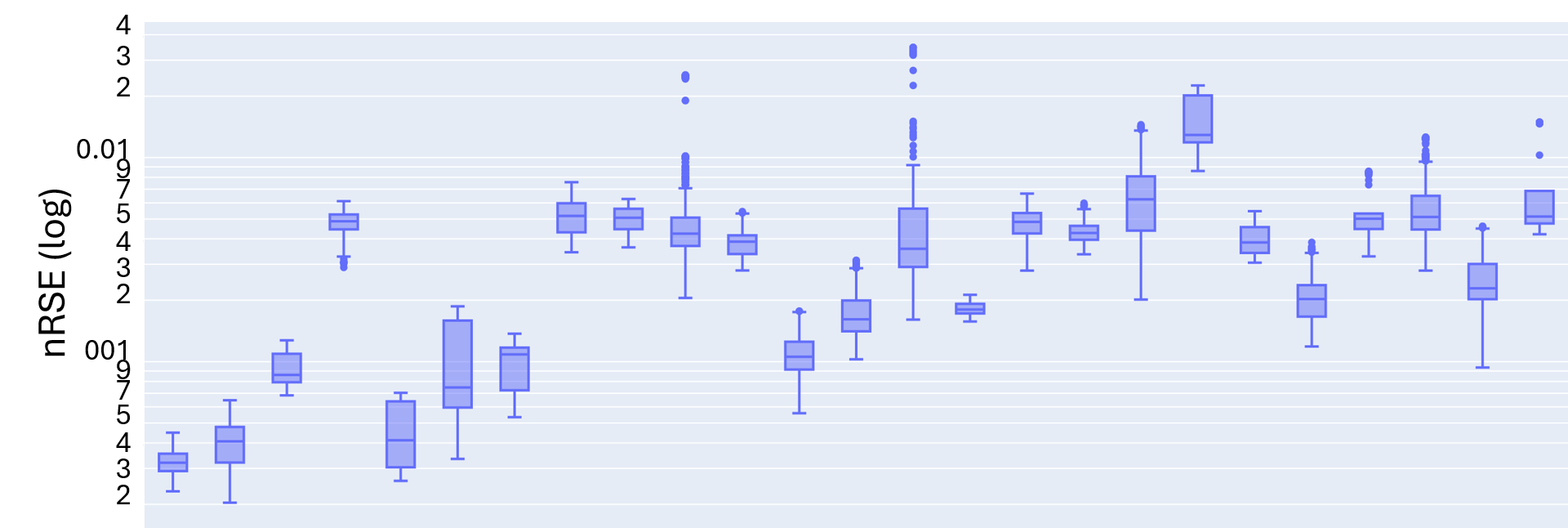}}
	\caption{Per-building evaluation of \gls{nrse} distributions for 24-hour prediction, logarithmically scaled. The series are sorted by average energy consumption, ascending from left to right.}
	\label{fig:nrse-per-building-24-hours}
\end{figure}

The \gls{xlstm} for 24-hour prediction achieves a \gls{rmse} of 21.47 \gls{kwh}, the value is averaged over all test series in the dataset. The per-building analysis is visualized in Fig. \ref{fig:nrse-per-building-24-hours}, where the distribution of the \gls{nrse} is drawn for each series in the test dataset individually. The \gls{nrse} is calculated by normalizing the \gls{rse} with the average consumption of the building in the test set. For the 24-hour prediction, one can see that the distributions of the \gls{nrse} are heterogeneous, with no immediate correlation between increasing size of the building (average consumption) and \gls{nrse} distributions.

\subsection{Time and Memory Complexity Comparison}\label{sec:time-and-memory-complexity-comparison}

Important factors of sustainable computing are model size and training efficiency, which directly influence energy consumption, carbon emissions, and resource utilization. \glspl{nn} vary significantly in their computational demands, making their time and memory complexity crucial for model selection in heat load forecasting applications. This section compares the computational requirements of the tested models for 24-hour prediction to provide insights for resource-effective practical implementation.

\subsubsection{Memory Requirements}
\glspl{fcn} require memory proportional to the number of weights between layers. Our implementation, with only one layer with 131 neurons, demonstrated remarkable learnable parameter efficiency (8,277 trainable parameters).
\glspl{lstm} have a memory complexity of $O(nd)$, for sequence length $n$ and embedding dimension $d$. Our \gls{lstm} implementation sums to 71,296 trainable parameters.
The \gls{xlstm} architecture increases memory usage by replacing scalar hidden states with matrix representations of size $d \times d$. Therefore, the \gls{xlstm} requires about 2.1 million learnable parameters.

The next larger model is the Transformer, with about 3.7 million learnable parameters. In general, Transformers have the highest memory requirements, scaling quadratically with input sequence length due to the $O(n^2)$ attention matrices. A vanilla Transformer requires approximately $O(4Ld^2 + 2Ld)$ parameters across $L$ layers. The \gls{tft} introduces additional memory requirements for their specialized components, which causes its size to grow to about 5.7 million trainable parameters. This is the biggest model of the benchmark.

\subsubsection{Computational Complexity}

The models are trained on \glspl{cpu} and \glspl{gpu}, depending on their complexity and compatibility. The TFT could only be trained on the \gls{cpu}, because the provided TFT code runs on TensorFlow version 1.15, which was incompatible with available GPUs. The \gls{cpu} that is used is an Intel Core i7-1165G7. The \gls{gpu} used is a NVIDIA L40S.

\begin{table}[htbp]
\centering
\caption{Runtime and Energy Metrics for best model trainings for 24-hour prediction. Trainings on GPU are marked with a *.}
\label{tab:runtime_hourly_24}
\begin{tabular}{lccccc}
\toprule
Metric & xLSTM*& LSTM* & FCN & TE & TFT\\
\midrule
\makecell[l]{Runtime per \\Epoch (min)} & 3.64 & 0.51 & 0.02 & 0.29 & 41\\
Epochs & 30 & 49 & 73 & 10 & 20\\
\makecell[l]{Total CO$_2$\\Emissions (gCO$_2$e)} & 34.17 & 6.47 & 0.18 & 0.79 & 38.22\\
\bottomrule
\end{tabular}
\end{table}

Training times varied dramatically, with \glspl{fcn} completing training in under two minutes across both forecasting horizons. The \gls{xlstm} models, when trained on GPU, required 10-20 minutes, while \glspl{tft} demanded 6-11 hours for hourly predictions on CPU. \glspl{lstm} showed moderate performance, with training times of about an hour on the CPU and 10 minutes on the GPU. An overview of runtimes and total CO$_2$ emissions generated while training the best models for 24-hour prediction can be found in Tab. \ref{tab:runtime_hourly_24}. The energy metric analysis reveals a significant trade-off: \gls{xlstm}'s superior forecasting accuracy comes at a substantial environmental cost, producing approximately 190 times more CO$_2$ emissions compared to the \gls{fcn}. The \gls{tft} training on CPU exhibits the highest emission of CO$_2$, due to its long runtime. For practical applications, these findings suggest that simpler models like \glspl{fcn} offer an excellent balance of efficiency and performance when computational resources are limited. More complex models may be justified when forecast accuracy is paramount and computational resources are abundant.

\section{Discussion}\label{sec:discussion}

All tested models demonstrated the ability to outperform the na$\ddot{\i}$ve forecast across various experiments, confirming their capacity to learn from heterogeneous building datasets. The baseline \gls{fcn} model stands out for its efficiency in both time and memory usage while achieving competitive results.

The \gls{xlstm} model demonstrated superior performance in both experiments, showing remarkable ability to capture patterns. Despite not being trained with future covariates, it competed effectively against models that utilized this additional information. Its performance could potentially improve if the architecture were adapted to incorporate future covariates.

The Transformer-based models (\gls{te} and \gls{tft}) presented contrasting results. While the \gls{te} generally underperformed despite its size and training requirements, the \gls{tft} achieved the best \gls{nrmse} values for three-hour predictions, as well as the lowest \gls{mae}. However, these models required a significantly larger amount of trainable parameters to attain those scores.

The \gls{xlstm} consistently performed well on \gls{rmse} metrics, but was often outperformed on the \gls{mae}. This suggests that it effectively minimizes large errors, while making smaller consistent ones throughout predictions. 

A notable limitation in our experimental setup was the varying hyperparameter optimization across models. Smaller models like \gls{fcn} and \gls{lstm} benefited from extensive hyperparameter searches due to their short training times, while larger models like the \gls{te}, \gls{tft}, and \gls{xlstm} relied on theoretical optimal configurations. Additionally, technical constraints prevented \gls{gpu} acceleration for the \gls{tft} model, limiting exploration of its full potential. This mismatch resulted in slower training, which might have reduced the model's ability to reach an equally optimized solution compared to GPU‑trained architectures. As a result, the TFT's final accuracy may partially reflect hardware constraints rather than model capability. Likewise, training‑time comparisons across models are not strictly hardware‑fair, since other architectures were GPU‑accelerated. Future work should address this by porting the TFT to a modern TensorFlow or PyTorch implementation enabling GPU training, and by re-running experiments under consistent hardware conditions, either fully GPU‑based or fully CPU‑based.

Compared to recent research in energy forecasting, an \gls{rmse} of about 21.5 kWh for 24-hour prediction is not exceptional. For example, Eseye et al. \cite{eseye_short-term_2020} achieve an RMSE of 14.04 \gls{kwh} with their EMD-ICA-SVM model for 24-hour prediction of district heating of a residential building. It needs to be considered that our approach features a multi-building dataset, which has not been done by related works yet. Additionally, the dataset presented other unique challenges, like building time series varying in length and recording periods. While this complicated the forecasting task, it also created a more realistic scenario for practical applications in sustainable energy management systems. It demonstrates the usage of one model for forecasting the heat load of multiple buildings, reducing costs and energy compared to training and running multiple models separately. This is especially interesting when considering the scope of a city. Other works have focused on directly forecasting city-level consumption \cite{potocnik_machine-learning-based_2021, tang_modeling_2014, xue_multi-step_2019, kato_heat_2008, JOHANSSON2017208}. With our approach, the model could be trained on residential building level data, and finally be used to estimate the power plant level heat load for the next day, incorporating knowledge from each single building. This estimation can be used by the energy providers for efficient resource planning. At the same time, new buildings can benefit from a pre-trained model, that can be used to estimate their heat load, even when not being able to provide own train data. However, our current time‑split experiments only test forecasting for known buildings. To fully validate such cold‑start use cases, additional experiments with a cross‑building split (e.g., holding out entire buildings during training) are needed to assess whether the model can generalize reliably to unseen buildings.

\section{Conclusion}\label{sec:conclusion}
The experiments conducted in this study establish a comprehensive multi-building benchmark for heat-demand forecasting, demonstrating that models trained on pooled, heterogeneous building data can effectively generalize across both residential and commercial buildings. This is particularly significant as it addresses the challenge of developing forecasting systems that can work effectively across diverse building portfolios without requiring custom modeling for each structure.

Our head-to-head comparison between \gls{xlstm} and Transformer architectures against established baselines (\gls{fcn}, \gls{lstm}) using standardized inputs (history, weather, calendar, and building features) yielded insightful results. The \gls{xlstm} emerged as the top performer, achieving the lowest \gls{rmse} scores for both three-hour and 24-hour prediction horizons. The \gls{tft} exhibited superior MAE performance across both forecasting tasks, possibly related to its specialized feature processing architecture.

From a computational sustainability perspective, our assessment of training time and memory/parameter complexity revealed significant differences between the models that directly impact their operational carbon footprint and deployment feasibility. The \gls{fcn} stood out as the most resource-efficient model, with the least amount of trainable parameters and fastest training time, making it ideal for hardware-constrained environments and edge deployment scenarios. In contrast, the \gls{tft} had the largest amount of parameters, requiring substantially more computational resources. The \gls{xlstm} offered an attractive middle ground, delivering top performance with moderate computational demands. These findings highlight the importance of considering not only prediction accuracy but also computational efficiency when selecting models for real-world energy forecasting applications, thereby aligning forecasting performance with broader climate-sustainability goals.

\clearpage
\bibliographystyle{ieeetr}
\bibliography{LiteratureSusTech2026}

\begin{thebibliography}{10}

\bibitem{energy_and_climate}
{German Federal Ministry for Economic Cooperation and Development}, ``Energy
  and climate.''
  {https://www.bmz.de/en/issues/climate-change-and-development/energy-and-climate},
  2025.

\bibitem{energy_transition}
T.~Pan, ``Why heat is a challenge in the fight against climate change, and what
  we can do about it,'' 2023.

\bibitem{djebko_design_2024}
K.~Djebko, D.~Weidner, M.~Waleska, T.~Krey, B.~Kamble, S.~Rausch, D.~Seipel,
  and F.~Puppe, ``Design and implementation of a decision integration system
  for monitoring and optimizing heating systems: Results and lessons learned,''
  {\em Energies}, vol.~17, no.~24, p.~6290, 2024.

\bibitem{EnergyInformatics2020}
A.~Vandermeulen, B.~van~der Heijde, D.~Patteeuw, D.~Vanhoudt, and L.~Helsen,
  ``An operational strategy for district heating networks: application of
  energetic flexibility,'' {\em Energy Informatics}, vol.~3, no.~1, pp.~1--17,
  2020.

\bibitem{CarbonPeak2021}
Y.~Jiang, Q.~Wan, X.~Yang, Y.~Xu, and Y.~Wang, ``Cooling, heating and electric
  load forecasting for integrated energy system based on lstm-cnn,'' in {\em
  2021 IEEE 4th International Electrical and Energy Conference (CIEEC)},
  pp.~1--6, IEEE, 2021.

\bibitem{Feng2024}
Y.~Feng, ``Optimizing energy efficiency: predicting heating load with a machine
  learning approach and meta-heuristic algorithms,'' {\em Multiscale and
  Multidisciplinary Modeling, Experiments and Design}, vol.~7, pp.~3993--4009,
  2024.

\bibitem{zhu2023time}
J.~Zhu, Z.~Zhao, X.~Zheng, Z.~An, Q.~Guo, Z.~Li, J.~Sun, and Y.~Guo,
  ``Time-series power forecasting for wind and solar energy based on the
  sl-transformer,'' {\em Energies}, vol.~16, no.~22, p.~7610, 2023.

\bibitem{bayer2024electricity}
D.~R. Bayer, F.~Haag, M.~Pruckner, and K.~Hopf, ``Electricity demand
  forecasting in future grid states: A digital twin-based simulation study,''
  in {\em 2024 9th International Conference on Smart and Sustainable
  Technologies (SpliTech)}, pp.~1--6, IEEE, 2024.

\bibitem{sherstinsky_fundamentals_2023}
A.~Sherstinsky, ``Fundamentals of recurrent neural network ({RNN}) and long
  short-term memory ({LSTM}) network,'' {\em Physica D: Nonlinear Phenomena},
  vol.~404, p.~132306, 2020.

\bibitem{vaswani_attention_nodate}
A.~Vaswani, N.~Shazeer, N.~Parmar, J.~Uszkoreit, L.~Jones, A.~N. Gomez,
  {\L}.~Kaiser, and I.~Polosukhin, ``Attention is all you need,'' {\em Advances
  in neural information processing systems}, vol.~30, 2017.

\bibitem{zhou_informer_2021}
H.~Zhou, S.~Zhang, J.~Peng, S.~Zhang, J.~Li, H.~Xiong, and W.~Zhang,
  ``Informer: Beyond efficient transformer for long sequence time-series
  forecasting,'' in {\em Proceedings of the AAAI conference on artificial
  intelligence}, vol.~35, pp.~11106--11115, 2021.

\bibitem{beck2024xlstmextendedlongshortterm}
M.~Beck, K.~P{\"o}ppel, M.~Spanring, A.~Auer, O.~Prudnikova, M.~Kopp,
  G.~Klambauer, J.~Brandstetter, and S.~Hochreiter, ``xlstm: Extended long
  short-term memory,'' {\em Advances in Neural Information Processing Systems},
  vol.~37, pp.~107547--107603, 2024.

\bibitem{tay2021long}
Y.~Tay, M.~Dehghani, S.~Abnar, Y.~Shen, D.~Bahri, P.~Pham, J.~Rao, L.~Yang,
  S.~Ruder, and D.~Metzler, ``Long range arena : A benchmark for efficient
  transformers,'' in {\em International Conference on Learning
  Representations}, 2021.

\bibitem{hong_energy_2020}
T.~Hong, P.~Pinson, Y.~Wang, R.~Weron, D.~Yang, and H.~Zareipour, ``Energy
  {Forecasting}: {A} {Review} and {Outlook},'' {\em IEEE Open Access Journal of
  Power and Energy}, vol.~7, pp.~376--388, 2020.

\bibitem{eseye_short-term_2020}
A.~T. Eseye and M.~Lehtonen, ``Short-{Term} {Forecasting} of {Heat} {Demand} of
  {Buildings} for {Efficient} and {Optimal} {Energy} {Management} {Based} on
  {Integrated} {Machine} {Learning} {Models},'' {\em IEEE Transactions on
  Industrial Informatics}, vol.~16, pp.~7743--7755, Dec. 2020.

\bibitem{potocnik_machine-learning-based_2021}
P.~Poto\v{c}nik, P.~\v{S}kerl, and E.~Govekar, ``Machine-learning-based
  multi-step heat demand forecasting in a district heating system,'' {\em
  Energy and Buildings}, vol.~233, p.~110673, Feb. 2021.

\bibitem{tang_modeling_2014}
F.~Tang, A.~Kusiak, and X.~Wei, ``Modeling and short-term prediction of {HVAC}
  system with a clustering algorithm,'' {\em Energy and Buildings}, vol.~82,
  pp.~310--321, Oct. 2014.

\bibitem{kato_heat_2008}
K.~Kato, M.~Sakawa, K.~Ishimaru, S.~Ushiro, and T.~Shibano, ``Heat load
  prediction through recurrent neural network in district heating and cooling
  systems,'' in {\em 2008 {IEEE} {International} {Conference} on {Systems},
  {Man} and {Cybernetics}}, pp.~1401--1406, Oct. 2008.
\newblock ISSN: 1062-922X.

\bibitem{muzaffar_short-term_2019}
S.~Muzaffar and A.~Afshari, ``Short-{Term} {Load} {Forecasts} {Using} {LSTM}
  {Networks},'' {\em Energy Procedia}, vol.~158, pp.~2922--2927, Feb. 2019.

\bibitem{li_assessment_2021}
G.~Li, X.~Zhao, C.~Fan, X.~Fang, F.~Li, and Y.~Wu, ``Assessment of long
  short-term memory and its modifications for enhanced short-term building
  energy predictions,'' {\em Journal of Building Engineering}, vol.~43,
  p.~103182, Nov. 2021.

\bibitem{xue_multi-step_2019}
P.~Xue, Y.~Jiang, Z.~Zhou, X.~Chen, X.~Fang, and J.~Liu, ``Multi-step ahead
  forecasting of heat load in district heating systems using machine learning
  algorithms,'' {\em Energy}, vol.~188, p.~116085, Dec. 2019.

\bibitem{zerveas_transformer-based_2020}
G.~Zerveas, S.~Jayaraman, D.~Patel, A.~Bhamidipaty, and C.~Eickhoff, ``A
  transformer-based framework for multivariate time series representation
  learning,'' in {\em Proceedings of the 27th ACM SIGKDD Conference on
  Knowledge Discovery \& Data Mining}, KDD '21, (New York, NY, USA),
  p.~2114–2124, Association for Computing Machinery, 2021.

\bibitem{wu_autoformer_2021}
H.~Wu, J.~Xu, J.~Wang, and M.~Long, ``Autoformer: {Decomposition}
  {Transformers} with {Auto}-{Correlation} for {Long}-{Term} {Series}
  {Forecasting},'' in {\em Advances in {Neural} {Information} {Processing}
  {Systems}}, vol.~34, pp.~22419--22430, Curran Associates, Inc., 2021.

\bibitem{zhao_short-term_2021}
Z.~Zhao, C.~Xia, L.~Chi, X.~Chang, W.~Li, T.~Yang, and A.~Y. Zomaya,
  ``Short-{Term} {Load} {Forecasting} {Based} on the {Transformer} {Model},''
  {\em Information}, vol.~12, p.~516, Dec. 2021.
\newblock Number: 12 Publisher: Multidisciplinary Digital Publishing Institute.

\bibitem{zhang_short-term_2022}
G.~Zhang, C.~Wei, C.~Jing, and Y.~Wang, ``Short-{Term} {Electrical} {Load}
  {Forecasting} {Based} on {Time} {Augmented} {Transformer},'' {\em
  International Journal of Computational Intelligence Systems}, vol.~15, p.~67,
  Aug. 2022.

\bibitem{huy_short-term_2022}
P.~C. Huy, N.~Q. Minh, N.~D. Tien, and T.~T.~Q. Anh, ``Short-{Term}
  {Electricity} {Load} {Forecasting} {Based} on {Temporal} {Fusion}
  {Transformer} {Model},'' {\em IEEE Access}, vol.~10, pp.~106296--106304,
  2022.

\bibitem{giacomazzi2023short}
E.~Giacomazzi, F.~Haag, and K.~Hopf, ``Short-term electricity load forecasting
  using the temporal fusion transformer: Effect of grid hierarchies and data
  sources,'' in {\em Proceedings of the 14th ACM International Conference on
  Future Energy Systems}, pp.~353--360, 2023.

\bibitem{alharthi2024xlstmtime}
M.~Alharthi and A.~Mahmood, ``xlstmtime: Long-term time series forecasting with
  xlstm,'' {\em AI}, vol.~5, no.~3, pp.~1482--1495, 2024.

\bibitem{kraus_xlstm-mixer_2024}
M.~Kraus, F.~Divo, D.~S. Dhami, and K.~Kersting, ``{xLSTM}-mixer: Multivariate
  time series forecasting by mixing via scalar memories,'' 2024.

\bibitem{hochreiter1997long}
S.~Hochreiter and J.~Schmidhuber, ``Long short-term memory,'' {\em Neural
  computation}, vol.~9, no.~8, pp.~1735--1780, 1997.

\bibitem{lim2020temporalfusiontransformersinterpretable}
B.~Lim, S.~O. Arık, N.~Loeff, and T.~Pfister, ``Temporal fusion transformers
  for interpretable multi-horizon time series forecasting,'' {\em International
  Journal of Forecasting}, vol.~37, no.~4, pp.~1748--1764, 2021.

\bibitem{ki_in_fernwaerme}
``Ai in district heating.''
  https://future-energy-lab.de/projects/ki-in-fernwaerme/.
\newblock Accessed: 23.05.2025.

\bibitem{bourdeau_modeling_2019}
M.~Bourdeau, X.~Q. Zhai, E.~Nefzaoui, X.~Guo, and P.~Chatellier, ``Modeling and
  forecasting building energy consumption: {A} review of data-driven
  techniques,'' {\em Sustainable Cities and Society}, vol.~48, p.~101533, July
  2019.

\bibitem{kingma2014adam}
D.~P. Kingma and J.~Ba, ``Adam: A method for stochastic optimization,'' {\em
  arXiv preprint arXiv:1412.6980}, 2014.

\bibitem{glorot2010understanding}
X.~Glorot and Y.~Bengio, ``Understanding the difficulty of training deep
  feedforward neural networks,'' in {\em Proceedings of the thirteenth
  international conference on artificial intelligence and statistics},
  pp.~249--256, JMLR Workshop and Conference Proceedings, 2010.

\bibitem{JOHANSSON2017208}
C.~Johansson, M.~Bergkvist, D.~Geysen, O.~D. Somer, N.~Lavesson, and
  D.~Vanhoudt, ``Operational demand forecasting in district heating systems
  using ensembles of online machine learning algorithms,'' {\em Energy
  Procedia}, vol.~116, pp.~208--216, 2017.
\newblock 15th International Symposium on District Heating and Cooling,
  DHC15-2016, 4-7 September 2016, Seoul, South Korea.

\end{thebibliography}

\end{document}